\documentclass[letterpaper, 10pt, conference]{ieeeconf}      
\IEEEoverridecommandlockouts
\usepackage[centertags]{amsmath}
\usepackage{graphics}
\usepackage{amscd}
\usepackage{amsfonts}
\usepackage{amssymb}
\usepackage{helvet}
\usepackage{rotating}
\usepackage{epsfig}
\usepackage[english,ngerman]{babel}
\usepackage{babelbib} 	
\setbtxfallbacklanguage{english}
\usepackage[latin1]{inputenc}
\usepackage{placeins}
\usepackage[sort,nospace,compress]{cite} 
\usepackage{color}

\newcommand{\bE}{\mathbb{E}}

\newcommand{\bR}{\mathbb{R}}
\newcommand{\bS}{\mathbb{S}}
\newcommand{\bU}{\mathbb{U}}
\newcommand{\bV}{\mathbb{V}}
\newcommand{\bPf}{\mathbb{P}_{\text{\rm f}}}
\newcommand{\bPoc}{\mathbb{P}_{\text{\rm oc}}}
\newcommand{\bPnd}{\mathbb{P}_{\text{\rm nd}}}
\newcommand{\bC}{C}
\newcommand{\OO}{O}
\newcommand{\OOU}{O_u}
\newcommand{\OOS}{B^{r}_s}
\newcommand{\OOD}{B^{r}_d}
\newcommand{\cP}{\mathcal{P}}
\newcommand{\cN}{\mathcal{N}}
\newcommand{\Model}{M}

\def\equals{\mathrel{\widehat{=}}}

\usepackage[pdftex,
	bookmarks=true,
	bookmarksnumbered=true,
	hypertexnames=true,
	breaklinks=true,
	plainpages=false,
	colorlinks=true,
	linkcolor=black,
	citecolor=black,
	urlcolor=blue]{hyperref}
\usepackage{fancyhdr}
\renewcommand{\headheight}{12pt}

\definecolor{headercolor}{rgb}{.85,.85,.85}

\title{Optimal Camera Placement to measure Distances Conservativly Regarding Static and Dynamic Obstacles}

\author{ 
	\parbox{6.5cm}{\centering M. H\"anel and S. Kuhn and D. Henrich
	\thanks{ This work is part of the project SIMERO 2 and is supported by Deutsche Forschungsgemeinschaft.\newline
	J\"urgen Pannek was partially supported by DFG Grant Gr1569/12-1 within the priority research program 1305.}\\
	Angewandte Informatik III\\
	Universit\"at Bayreuth\\
	95440 Bayreuth, Germany\\
	{\tt\small \{maria.haenel,stefan.kuhn,\\
	dominik.henrich\}@uni-bayreuth.de}}
	\hspace*{ 0.5 in}
	\parbox{6.5cm}{ \centering J. Pannek and L. Gr\"une\\
	Lehrstuhl f\"ur Angewandte Mathematik\\
	Universit\"at Bayreuth\\
	95440 Bayreuth, Germany\\
	{\tt\small \{juergen.pannek,lars.gruene\}\\
	@uni-bayreuth.de}}
}

\date{} 

\begin{document}
\maketitle

\begin{abstract}
In modern production facilities industrial robots and humans are supposed to interact sharing a common working area. In order to avoid collisions, the distances between objects need to be measured conservatively which can be done by a camera network. To estimate the acquired distance, unmodelled objects, e.g., an interacting human,  need to be modelled and distinguished from premodelled objects like workbenches or robots by image processing such as the background subtraction method.

The quality of such an approach massively depends on the settings of the camera network, that is the positions and orientations of the individual cameras. Of particular interest in this context is the minimization of the error of the distance using the objects modelled by the background subtraction method instead of the real objects. Here, we show how this minimization can be formulated as an abstract optimization problem. Moreover, we state various aspects on the implementation as well as reasons for the selection of a suitable optimization method, analyze the complexity of the proposed method and present a basic version used for extensive experiments.
\end{abstract}


\section{Introduction}

Nowadays, human/machine interaction is no longer restricted to humans programming machines and operating them from outside their working range. Instead, one tries to increase the efficiency of such a cooperation by allowing both actors to share the same working area. In such a context, safety precautions need to be imposed to avoid collisions, i.e., the distance between human and machine interacting in a common area needs to be reconstructed continuously in order to detect critical situations. To this end, usually a network of cameras is installed to, e.g., ensure that every corner of the room can be watched, every trail can be followed or every object can be reconstructed correctly. Within this work, we focus on computing an optimal configuration of the camera network in order to measure the distances as correct as possible but still conservatively.

After a brief review on previous results concerning the predescribed distance measurement, we show how an unmodelled (human) object can be contoured by a 3D background subtraction method. We extend this scheme to cover both static and dynamic obstacles, some of which are modelled in advance but still occlude the vision of the sensor. In Section \ref{sec:visibility}, we rigorously formulate the problem of minimizing the error made by using the associated model instead of the original collective of  unmodelled objects. Considering the implementation of a solution method, we discuss various difficulties such as, e.g., the evaluation of the intersection of the cones corresponding to each camera in Section \ref{sec:discretizations} and also give an outline of concepts to work these issues. In the final Sections \ref{sec:experiments} and \ref{sec:conclusion}, we analyze the complexity of our basic implementation by a series of numerical experiments and conclude the article by given an outlook on methods to further improve the proposed method.

\section{State of the art}

Many camera placement methods have to deal with a trade-off between the quality of observations and the quantity of pieces of information which are captured by the cameras. The latter aspect is important for camera networks which have to decide \emph{whether} an item or an action has been observed. There have been investigations about how to position and orientate cameras subject to observing a maximal number of surfaces \cite{Holt2007} and different courses of action \cite{Bodor2007, Fiore2008, Fiore2008a} as well as maximizing the volume of the surveillance aread \cite{Murray2007} or the number of objects \cite{Mittal2004, Mittal2006, Mittal2008}. Another common goal in this context is to be able to observe all items of a given set but minimize the amount of cameras in addition to obtain their positions and orientations \cite{Erdem2004, Mittal2004, Mittal2006, Mittal2008}. This issue is called ``Art Gallery Problem'' especially when speaking of two--dimensional space. 

Apart from deciding whether an object has been detected by a camera network, another task is to obtain \emph{detailed} geometrical data of the observed item like its position and measurements of its corners, curves, surfaces, objects etc. As described in \cite{Luhmann2006} determining this information for distances smaller than a few hundred meters by cameras belongs to the field `close range photogrammetry'. 
In order to configure a camera network to cope with such tasks, one usually minimizes the error of observed and reconstructed items. Often the phrase `Photogrammetric Network Design' is used to express minimizing the reconstruction error for several (three-dimensional) points. The default assumption in this context, however, is that no occlusions occur, cf. \cite{Olague1998, Olague1998a, Olague2000, Olague2007,Hartley1992} for details. Optimally localizing an entire object which is not occluded is an assignment treated in \cite{Ercan2006a}. Furthermore, many approaches compensate for the increasing complexity of the problem by oversimplifying matters: One common approach is to restrain the amount of cameras (in \cite{Hartley1992} two cameras are used) or their position and orientation. Considering the latter, known approaches are the viewing sphere model given in \cite{Olague1998a, Olague1998} or the idea of situating all cameras on a plane and orientating them horizontally, cf. \cite{Ercan2006a}).

In contrast to these approaches, we discuss optimizing positions and orientations of cameras in a network in the context of the back\-ground sub\-traction method which is used to determine a visual hull of a solid object. By means of this visual hull, distances can be computed easily which renders this approach to be a different simplification. 
Occlusions of solids to be reconstructed obscure the view and enlarge the visual hull. In order to get the minimal error of the construction of the hull, \cite{Yang2004} assumes that minimizing the occuring occlusions of solids also reduces and thus specifies their possibile locations. However, neither obstacles nor opening angles other than $\pi$ are discussed in \cite{Yang2004} and additionally the orientation of the camera is neglected as a variable since it is simply orientated towards the object. In \cite{Ercan2006}, static obstacles are considered but the amount of cameras is chosen out of a preinstalled set. 

Since we are not interested in optimizing the quantity of observed objects but the quality of data, our approach is different to most of the discussed results. Note that the quality of information can be obtained by various types of image processing. Here, we consider the background subtraction method to obtain a visual hull of a given object. Within our approach, we optimize the positions and orientations of a fixed amount of cameras as to minimize the error that is made by evaluating distances to the visual hull. In contrast to existing results, our goal is to incorporate the aspects of occuring static or dynamic obstacles into our calculations but also to exploit all degrees of freedom available in an unconstrained camera network. Nevertheless, distances are to be evaluated conservativly.

\section{Visibility Analysis}
\label{sec:visibility}


Within this section, it will be shown how to condition the objective function on the cameras position and orientation, thus, we successively build up a mathematical representation of the optimization problem. We start off by defining the critical area as well as the to be reconstructed unmodelled objects and corresponding abstract models in Section \ref{subsec:formalizing}. This will allow us to formally state the objective function which we aim to minimize. In the following Section \ref{subsec:model}, we define the camera network and its degrees of freedom, i.e. the position and orientation of each camera. These degrees of freedom allow us to parametrize the model of the to be reconstructed object. Additionally, this tuple of degrees of freedom will serve as an optimization variable in the minimization problem stated in Section \ref{subsec:obstacles}. To cover all possible scenarios, this problem is extended by incorporating both static and dynamic obstacles as well as an evolving time component.

\subsection{Formalizing the Problem}
\label{subsec:formalizing}

Let $\bU \subset \bR^3$ be a spacial area based on which information about humans, perils, obstacles and also cameras can be given. 
Consider $\bS \subset \bU$ to be the surveillance area, where critical points of the set $\bC \subset \bS$ as well as objects, such a human or a robot, are monitored. 

For the moment, we neglect obstacles completely, we just distinguish two types of objects, to explain the basic idea of reconstructing an object by the means of a camera network: 
If a detailed model of an object exists describing its appearance like location, shape, color or else, the object is called \emph{modelled}. If this is not the case the object is called \emph{unmodelled}.
This is motivated by the following scenario: If humans move unpredictably within the surveillance area, i.e. without a given route, their appearance is unmodelled and needs to be reconstructed to be used for further calculations. 
The model of an unmodelled object can be reconstructed by the means of a camera network.
Therefore, let $\OOU(a) \subset \bS$ be a complete set of points included in one or more unmodelled objects, depending on the appearance of unmodelled objects specified by the parameter $a\in \bR^k$. 
We refer to these objects as \emph{unmodelled collective}.
Since automaticly placing the cameras for such a scenario is incomputable without information on the unmodelled collective, we impose the assumption that the distribution $\cP: 2^{\bR^k} \to [0,1]$ of the appearance $a \in \bR^k$ is known. 

\begin{figure}[!ht]\centering
	\begin{minipage}[h]{0.85\linewidth}\centering
	\includegraphics[width=0.8\linewidth] {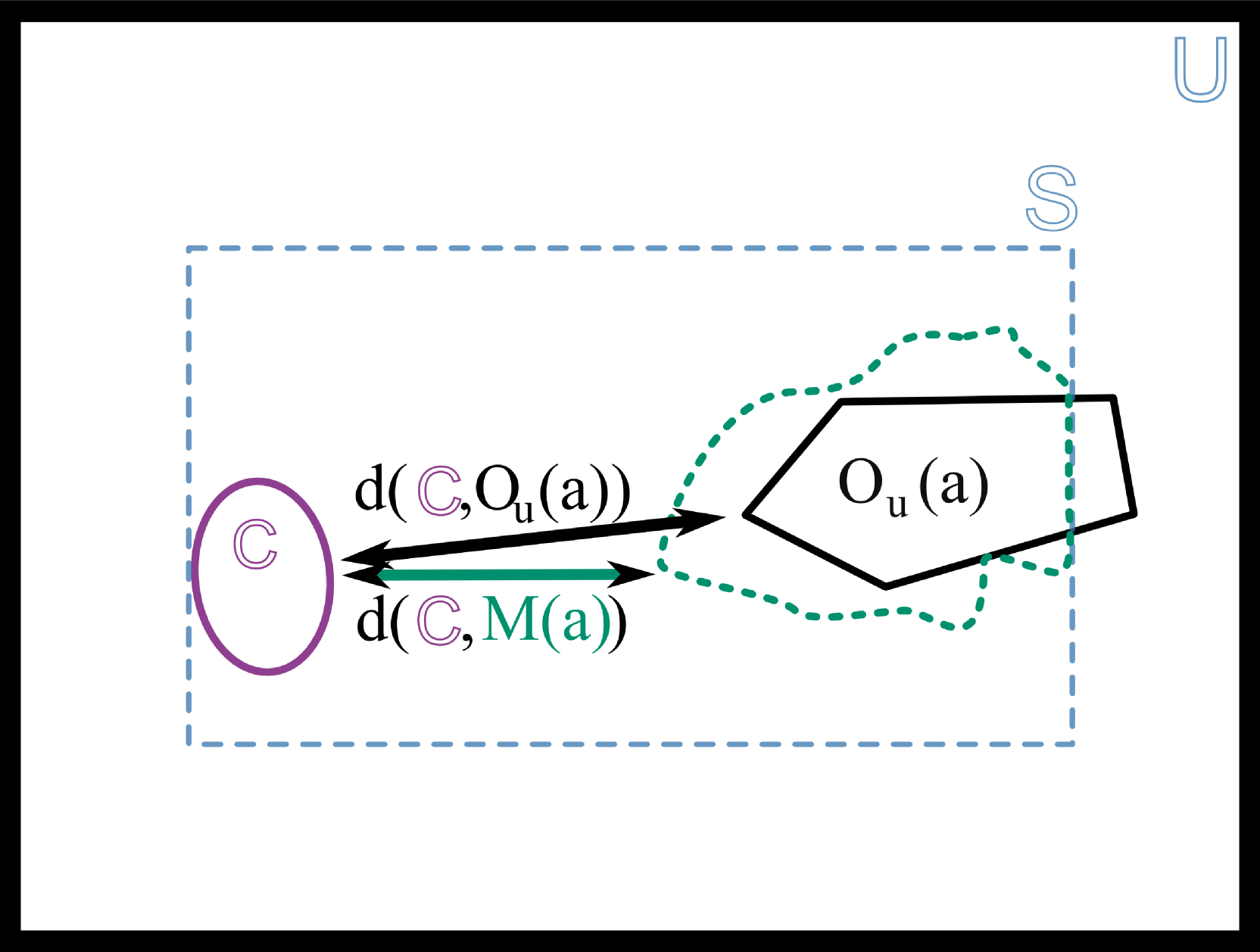}
	\caption{Surveillance area $\bS$: Distance between critical points $\bC$ and unmodelled collective (black) and distance to the approximated model (green)}
	\label{pic:distance}
	\end{minipage}
\end{figure}

As the safety of a human being must be guaranteed in any case, the distance
\begin{align*}
	d(\bC, \OOU(a)) := \min \{ d(x, y) \mid x \in \bC, y \in \OOU(a) \}
\end{align*}
has to be computed conservatively and security measures need to be taken if the unmodelled collective $\OOU(a)$ appoaches the critical points $\bC$. Here $d(\cdot, \cdot)$ denotes a standard distance function. 
If the exact set $\OOU(a)$ was known, this distance could be evaluated easily. 
As we do not directly know the value of $a \in \bR^k$ and therefore can only guess the points that are included in $\OOU(a)$, we need to approximate a (as a consequence also conservative) model $\Model(a)\subset\bS$, see Fig. \ref{pic:distance}.

Note that for now $\Model(a)$ is an abstract approximation of $\OOU(a)$ with respect to the parameter $a$ only. In order to actually compute $\Model(a)$, a sensor network and its degrees of freedom come into play, see Section \ref{subsec:model} for details. Still, the abstract approximation allows us to formalize our overall task, i.e. to minimize the difference between the approximation based distance $d(\bC,\Model(a))$ and the real distance $d(\bC, \OOU(a)))$. Taking the assumed distribution of the parameter $a$ into account, we aim to minimize the functional
\begin{align}
	\label{eq:error}
	\int\limits_{a \in \bR^k} \Big[ d(\bC, \OOU(a)) - d(\bC, \Model(a))\Big]^2\ d\cP(a).
\end{align}

Note that for the optimization we need to be aware of possible appearances of the object in order to let the integral pass through their space $a\in\bR^k$. Thus all appearances $a\in\bR^k$ of the unmodelled collective should be known.

\subsection{Building a Model with the Camera Network}
\label{subsec:model}

In the previous section we saw that in order to evaluate the functional \eqref{eq:error}, a model $\Model(a)$ of the unmodelled collective $\OOU(a)\subset\bU$ is required. To obtain such a model, we impose a camera network $\cN$ consisting of $n\in\mathbb{N}$ Cameras. Each camera can be placed and orientated with a setting $\bE = (\bU \times [-\pi,\pi] \times [-\frac{\pi}2, \frac{\pi}2])$. Here, the first term corresponds to the position of the camera whereas the second and third denote the angles `yaw' and `pitch' respectively. For simplicity of exposition, we exclusively considered circular cones in our implementation which allowed us to neglect the angle `roll' as a degree of freedom in the setting of a single camera. Hence, each camera exhibits five degrees of freedom, three for the \textit{position} and two for its \textit{orientation}.

Thus, each camera can be regarded as a tupel $(e, p) \in \bE \times \bU$ whereas its produced output regarding the parameter $a \in \bR^k$ of unmodelled collective is a function
\begin{alignat*}{10}
\kappa: \; & ( \bE & \times & \bU & \times & \bR^k) & & \to \bV & \\
	&( e, & & p, & & a) & & \mapsto \kappa_{e,a}(p) &
\end{alignat*}
that is - given the setting $e \in E$ and the appearance of the unmodelled collective $a \in \bR^k$ - each point $p \in \bU$ is mapped onto a sensor value $v \in \bV$ where
\begin{align*}
	\bV := \{\text{free}, \text{occupied}, \text{undetectable}\}.
\end{align*}

This set is adjusted to the evaluation of the network's images by the change detection method (e.g. background subtraction). The sensor value $\kappa_{e,a}(p)$ of a point $p\in\bU$ is \emph{free} if this point is perceived as not part of the unmodelled collective. The value \emph{occupied} resembles the possibility that the point could be part of the unmodelled collective (i.e. the point might be occupied by the collective). If the sensor cannot make the decision, e.g. this is the case for cameras that cannot `see' behind walls, the value is \emph{undetectable}. 
Obstacles like walls will be discussed in Section \ref{subsec:obstacles}. To obtain the values of set $\bV$ one could apply the method of background subtraction, which is discussed in \cite{kuhn2009} elaborately. 
Although our method is not restricted to a pixel model which is considered in \cite{kuhn2009}, the idea of this work remains the same. Thus, we will only provide the prior formulization of the values, as to explain their role in building the model of an unmodelled collective.

According to the definition of the set $\bV$, all cameras split the set $\bU$ into three different subsets:
\begin{align*}
	\bPf(e,a)  & = \{u\in\bU \mid \kappa_{e,a}(u) \equals \text{`free'}\} \\
	\bPoc(e,a) & = \{u\in\bU \mid \kappa_{e,a}(u) \equals \text{`occupied'}\} \\
	\bPnd(e,a) & = \{u\in\bU \mid \kappa_{e,a}(u) \equals \text{`undetectable'}\} 
\end{align*}
We state here without proof that we have constructed these parts to be a pairwise disjoint conjunction of $\bU$, i.e.
\begin{align*}
	\bU & = \bPf(e,a) \cup \bPoc(e,a) \cup \bPnd(e,a)
\end{align*}
with $\bPf(e,a) \cap \bPoc(e,a) = \bPf(e,a) \cap \bPnd(e,a) = \bPoc(e,a) \cap \bPnd(e,a) = \emptyset$ hold.

The unmodelled collective $\OOU(a)$ cannot be situated inside $\bPf(e,a)$, all we know is
\begin{align*}
	\OOU(a)\ \subset\ \bPoc(e,a) \cup \bPnd(e,a) \ =\ \bU\backslash \bPf(e,a).
\end{align*}
Since this inclusion holds for the parameter $a \in \bR^k$ and one camera with settings $e \in \bE$, obviously the following is true if we consider a camera network $\cN$ consisting of $n$ cameras with settings $e_i,\ i=1, \ldots,n$:
\begin{align*}
	\OOU(a) &\ \subset\ (\bU \backslash \bPf(e_1,a) \cap \ldots \cap \bU \backslash \bPf(e_n,a))\\
	&\ =\ \bU \backslash \big( \bPf(e_1,a) \cup \ldots \cup \bPf(e_n,a) \big)
\end{align*}
Note that this set is already a good approximation of the unmodelled collective if we considered the entire set $\bU$. However, as we only monitor the surveillance area $\bS$, we define the desired model $\Model(a)$ of the unmodelled collective $\OOU(a)$ as the intersection with the set $\bS$, i.e.,
\begin{align}
	\Model(a) & \equiv \Model(a, e_1, \ldots, e_n) \nonumber\\
	& := \bS \cap \Big( \bU \backslash \big( \bPf(e_1,a) \cup \ldots \cup \bPf(e_n,a) \big) \Big) 
\label{eq:model}
\end{align}
This is the basic model that can be used to calculate Formula \eqref{eq:error}. In the following, we will extend our setting to incorporate a time dependency and to cover for different types of obstacles.

\subsection{Adding Time and Obstacles}
\label{subsec:obstacles}

So far, we have only considered a static scene to be analyzed. 
Motivated by moving objects, we extend our setting by introducing a time dependency to the process under surveillance. 
Therefore, we declare the time interval of interest $I = [t_0, t_*]$, in which $t_0$ denotes the moment the reference image is taken and $t_*$ corresponds to the last instant the surveillance area ought to be observed.  
Thus, the unmodelled collective $\OOU(a(t))$, its probability distribution $\cP(a(t))$ and its approximation $\Model(a(t))\subset\bS$ as well as the set of critical points $\bC(t)$ change in time $t\in I$. 
As a simple extension of \eqref{eq:error} we obtain the time dependend error functional

\footnotesize\vspace*{-0.2cm}
	\begin{align}
		\label{eq:error_time}
		\int\limits_{t_0}^{t_*}\int\limits_{a(t)\in\bR^k} \Big[ d(\bC(t), \OOU(a(t))) - d(\bC(t), \Model(a(t),e_1, \ldots,e_n))\Big]^2\ d\cP(a(t)) dt
	\end{align}
\normalsize

In a second step, we add some more details to the scene under surveillance. To this end, we specify several categories and properties of objects $\OO\subset\bS$, which we are particularly interested in and which affect the reconstruction of the current scene.
Right from the beginning we have considered unmodelled objects. In contrast to modelled objects, these objects need to be reconstructed in order to track them. 
In the following, we additionally distinguish objects based on the characteristical behavior ``static/dynamic'', ``target/obstacle'' and ``rigid/nonrigid'', neglecting those objects that cannot be noticed by the sensors (like a closed glass door for cameras without distance sensor). 

We define a \emph{target} $T \subset \OO$ of a sensor network as an object which ought to be monitored and in our case reconstructed. 
An object $B\subset \OO$ which is not a target is called \emph{obstacle}. 
Furthermore, we distinguish obstacles based on their physical character: 
An obstacle $B$ features a \textit{rigid} nature (like furniture), if the inpenetrability condition $T \cap B^r = \emptyset$ holds, and is denoted by the index $r$ in $B^r$. 

The method proposed in \cite{kuhn2009} constructs a visual hull of an object by background subtraction, i.e. via change detection. In context of change detection methods another characteristical behavior of objects is relevant: 
A \emph{static object} is an object $\OO_s \subset \OO$ which is known to affect the given sensors in the same way at any time. 
If this is not the case, it is called \emph{dynamic}, which we indicate by adding a subscript and a time dependency $\OO_d(t)$.
More specifically, within the proposed background subtraction method the value of each pixel of a current image is subtracted from its counterpart within the reference image which has been taken beforehand. 
Thus, any change (like size/color/location) occuring after the reference image has been taken leaves a mark on the subtracted image, i.e., if the scene consists of static objects only, then the subtracted image is blank. 
For this reason, static objects must be placed in the scene before the reference picture is taken, and dynamic objects must not. 

Within the rest of this work, we consider all unmodelled objects to be reconstructed, i.e. in \eqref{eq:error_time} we have
\begin{align}
	\label{eq:target}
	\OOU(a(t)) := T(a(t))
\end{align}
Consequently, the unmodelled collective and its distance to the critical points are dynamic targets. 
Thus, we always consider an obstacle to be a \emph{modelled} obstacle since all our unmodelled objects are targets. Furthermore, all obstacles are considered rigid. 
To formalize the human-robot-scene let $\OOS \subset \OO$ and $\OOD(t) \subset \OO$ be the \emph{collective of static} and \emph{of dynamic obstacles} with time $t\in[t_0,t_*]$, respectively. 
We incorporate these new aspects into the model of the unmodelled collective in \eqref{eq:error_time} by intuitively extending our notation to
\begin{align}
	\label{eq:specific model}
	\Model(a(t),e_1, \ldots,e_n) := \Model(a(t),e_1, \ldots ,e_n,\OOS,\OOD(t)).
\end{align}
Last -- as a robot is a dynamic obstacle in addition to a security thread (f.e. when moving too fast) -- we define the critical points in \eqref{eq:error_time} as the collective of dynamic obstacles
\begin{align}
	\label{eq:critical points}
	\bC(t):=\OOD(t).
\end{align}

Note that there are dynamic obstacles next to dynamic targets i.e. the unmodelled collective. Thus a dynamical obstacle could easily be regarded as an object of the unmodelled collective since both evoke akin reactions of the change detection method. In our approach the obstacles are fully modelled and thus define a target free zone since they are physical obstacles. Still, inaccuracies of the accidental change detection leave fragments outside the dynamic obstacle, in our case outside the critical points. As a consequence the required distance between critical points and target is reduced to zero. Publication \cite{kuhn2009} solves this issue by introducing plausibility checks, in which predicates that characterize the target (like volume, height, etc.) are used to sort out the fragments.

In conclusion, our aim is to solve the problem
\begin{center}
	Minimize \eqref{eq:error_time}\\
	using definitions \eqref{eq:target}, \eqref{eq:specific model} and \eqref{eq:critical points}\\
	subject to $e_1, \ldots,e_n\in E$
\end{center}
i.e., to compute the optimal positions of $n$ cameras with settings $e_1, \ldots,e_n$ such that the measurement error is minimized.


 

\section{Aspects of Optimization}
\label{sec:discretizations}

There are various ways to compute Equation \eqref{eq:error_time} referring to: Representing the model, solving the integral and solving the optimization, as can be seen further on.

\subsection{Discretization of time and distribution}
\label{sec:discretizations_time}

\begin{figure}[!ht]\centering
	\begin{minipage}[h]{0.9\linewidth}\centering
 	\includegraphics[width=0.485\linewidth] {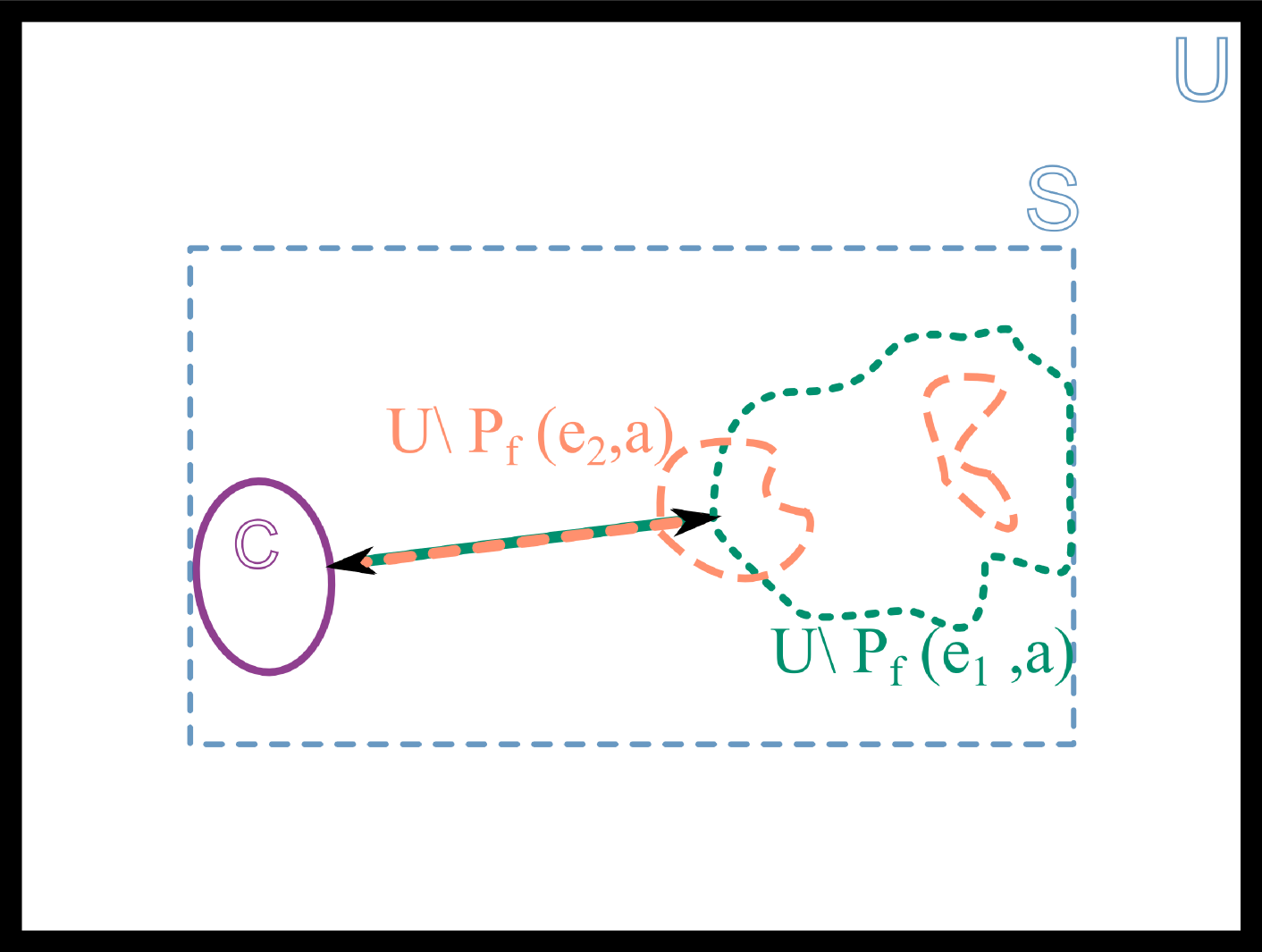}
 	\includegraphics[width=0.485\linewidth] {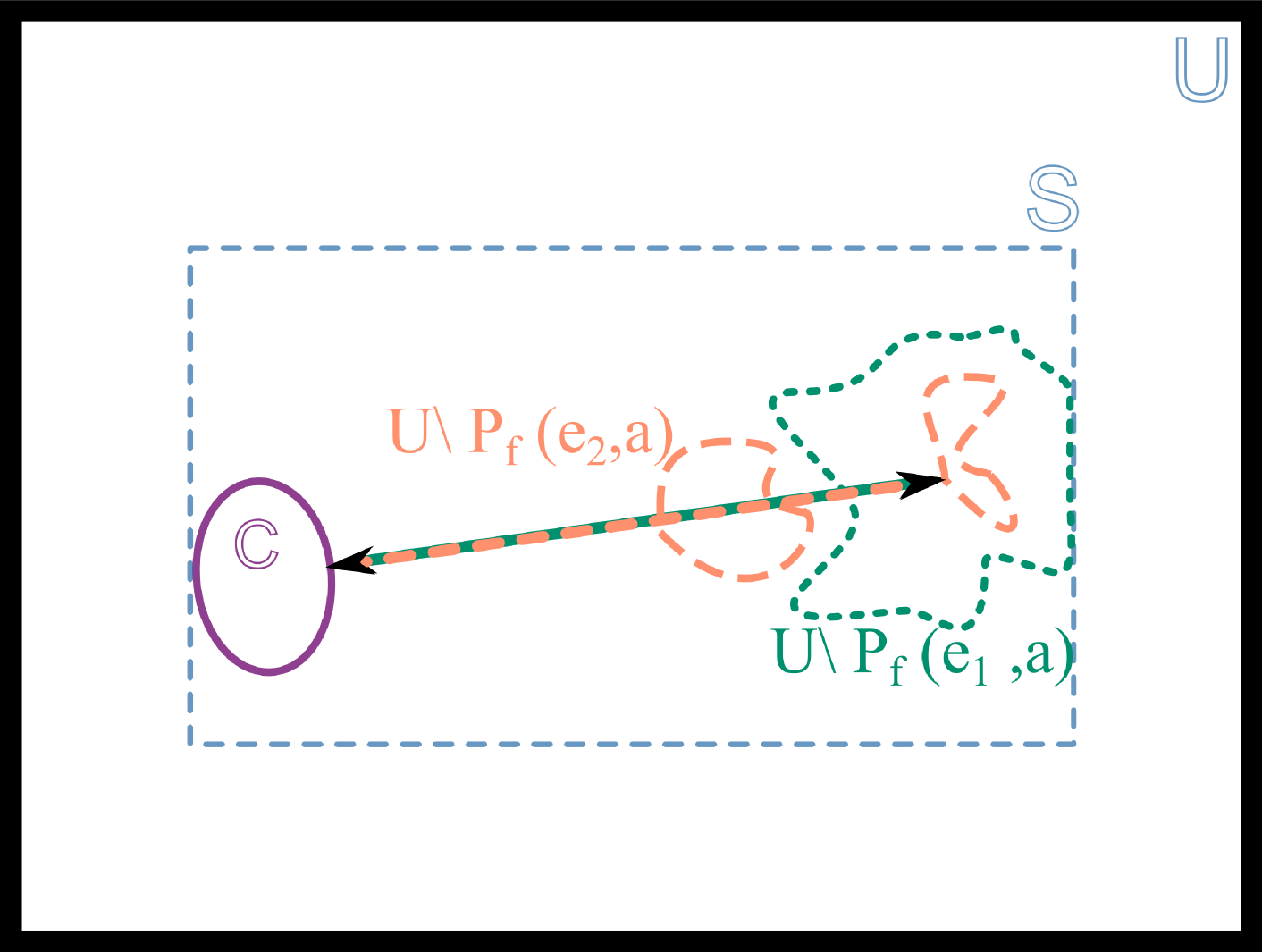}
	\caption{Discontinuity of the distance between perilous points $\bC$ and the approximated model, consisting of and intersection the nonfree part of camera 1, $\bU \setminus \bPf(e_1,a)$(green), and the nonfree part of camera 2, $\bPf(e_2,a)$(orange).}
	\label{pic:stetig}
	\end{minipage}
\end{figure}

We would at first like to state that the distance $d(\bC,\Model(a, \ldots3))$ between the model and another set does not need to be continuous at every appearance $a$ even if the distance $d(\bC,\OOU(a))$ to the unmodelled collective is continuous at $a$. This point can also be made for Equation \eqref{eq:error_time} but we stick to Equation \eqref{eq:error} for reasons of simplicity. Such a case is illustrated in Fig. \ref{pic:stetig}. As the original unmodelled collective $\OOU(a)$ of the appearance $a\in\bR^k$ does not necessarily need to be convex or even connected, given the settings $e_i, i=1, \ldots ,n\in E$, the unfree parts of the sensors $\bU\backslash \bPf(e_i,a)$ do not need to be connected, either.  The model is constructed of an intersection of these parts (see Equation \eqref{eq:model}). But, as intersections of disconnected parts do not need to be continuous on $a\in \bR^k$ (e.g. referring to Hausdorff--metrics), the distance $d(\bC,\Model(a,\ldots))$ between the model and another set does not need to be continuous at every appearance $a$.

Since only integrals with continuous integrants can generally be calculated as a whole or else need to be splitted, such a discontinuous function becomes a problem when being an integrant as of Formula \eqref{eq:error}. In our case a point of discontinuity of the distance as a function of $a$ cannot be derived easily, as it would have to be extracted from an individual nonrelated analysis depending not only on $a$ or $t$ but also on the sensor settings $e_i,i=1, \ldots,n$. While in simple cases this is possible, we spare such an altering analysis by discretizing appearance and time. Here, just the $l=1, \ldots,L$ most important appearances of the unmodelled collective $a_l\in \bR^k$ and $h=1, \ldots,H$ most important time steps $t_h\in[t_0,t_*]$ with $t_0=t_1$ and $t_*=t_H$ and with their weights $$\omega_{l,h}=\cP(a_l(t_h))\in[0,1]$$ are modelled. Accordingly, the following weighted sum approximates the integral of Formula \eqref{eq:error_time}:
\footnotesize
\begin{eqnarray}
Err_{L,H}(e_1, \ldots,e_n)&=&\\
	\sum_{h=1}^{H}\limits\sum_{l=1}^L\limits &\omega_{l,m}\cdot
			  &\Big[d\big(\bC(t_h),\OOU(a_l(t_h))\big) \nonumber\\
			&-&	d\big(\bC(t_h),\Model(a_l(t_h),e_1, \ldots,e_n,t_h)\big)\Big]^2 \nonumber
\label{eq:discretesum}
\end{eqnarray}
\normalsize

\subsection{Discretizing space by voxels}

The next challenge -- building an intersection of (free-form) solids -- has claimed to be subject of discussion for more than a quater of a century and still is an issue of recent investigations. The publication \cite{Wang2010} describes three main areas of solving this issue depending on their representation, each going with pros and cons. 

Solids represented by polygonial meshs can be intersected by exact arithmetic and intervall computation, checking surface membership afterwards. The major concerns of this approach are robustness and efficiency (e.g., while intersecting two tangetially connected polyhedra/polygons inside-out facettes are computed). 

Approximate methods (e.g. applying exact methods to a rough mesh of solids and refine the result) exist for meshs, too. Robustness problems (constructing breaks in the boundary) are in this case compensated by time consuming perturbation methods or interdependent operations which prevent parallel computing.

There are also techniques for solids transfered to image space (ray representation). While many of these mainly help rendering rather than evaluating the boundary, there are some that can be applied to intersection purposes (Layered Depth Image). Unfortunately, when computing these representations back into meshs many geometric details are destroyed.

Loosing geometric details is also the case for volumetric approaches. Converting surfaces with sharp corners and edges into volumetric data (like voxels) and not loosing data for reconstruction purpose is a challenging task even with oversampling. This also holds true for a voxel representation, but voxels on the other hand are easily obtained and robustly being checked by boolean operations. In addition to that we need a data structure, distances and volumes which are calculated easily, properties which are ensured for voxels. For these reasons our approach uses a voxel based model which is obtained by boolean operations on the free parts of the sensor.

\subsection{Optimization method}

After having evaluated existing solutions by plugging them in the objective function of a problem, the solver of an optimization problem is a strategy to improve solutions until an optimum of the objective function is reached. To choose a suitable solver for the specified problem, there are different characteristics of the objective function $Err(e_1, \ldots,e_n)$ that need to be considered. 

At first, we associate the cone of a camera subtracted from the surveillance area as the 'undetectable'-part of this camera, depending on the setting $e$ of the camera. Remember that the undetectable area could be part of the model of the unmodelled collective. Now imagine the cone rotating in 'yaw'-direction continuously. One can easily see that the distance between any given point of the surveillance area and this cone is not convex in $e$ (as an exception, the chosen point can be included in the 'undetectable'-part and the distance is therefore $0$ for all $e$). 

The second characteristic to be discussed is the discontinuity of $Err(e_1, \ldots, e_n)$ with respect to $e$. Due to the voxel based model distances are only evaluated to a finite set of points. When calculating the distances we need to jump from one point to the next even if settings are just altered gradually. Thus, the objective function is discontinuous and constant in between these discontinuities. Even if we used a non--voxelbased model, discontinuities would appear due to the intersections of disconnected parts mentioned in Section \ref{sec:discretizations_time}.

The objective function's properties complicate the search for a suitable solver. As elucidated in standard references on nonlinear optimization like \cite{NW2006}, most algorithms take advantage of a characteristical behavior like convexity, differentiability or at least continuity which cannot be guaranteed in our case. This applies to all determinisic solvers for nonlinear programs such as the Sequential Quadratic Programming, all kinds of local search algorithms (Downhill-Simplex, Bisection, Newton, Levenberg-Marquard etc.) and many others. Moreover, the problem cannot be transformed to a standard form of solvers like branch-and-bound, decompositions, cutting planes or outer approximation. This leaves us with non-deterministic, e.g., stochastic solvers. We have chosen the method MIDACO which is based on the ant-colony algorithm and samples solutions randomly where they appear to be most promising, see \cite{Schluter2010,Schluter2009a} for details.

\subsection{Complexity}






		


The solver is an iteration which generates a tuple of settings $e_i$, $i=1, \ldots ,n$ (one setting for each camera) within each iteration step stochastically, based on knowledge of previous generations. Given these settings the model, the distances and the objective function consisting of the weighted sum given in Equation \eqref{eq:discretesum} are evaluated. This continues until a stopping criteria is fulfilled. In order to compute the complexity of the method, assume that upon termination the $I$-th iteration step has been reached. The process of obtaining the objective value of Formula \eqref{eq:discretesum} is only implemented in a basic version, whereas for the given tuple of settings $e_i$ all of the $H$ time steps and $L$ appearances are to be evaluated to test all of the $r$ voxels whether they are included in the intersection in question. The intersection test uses all of the $f_u^{max}$ facets of the unmodelled collective as well as most of $f_s$ static facets and $f_d^{max}$ dynamic facets. Summing up these components give us the complexity
$$\mathcal{O}\Big(\ \mathnormal{I}\cdot r\cdot \left\{ \ n\  f_s\ +\ H\ (n + L)\ f_d^{max}\ +\ H\ L\ n\ f_u^{max}\ \right\}\ \Big)$$
of the method.

\section{Experiments}
\label{sec:experiments}


Since we use a stochastical solver on the non-convex problem of camera configuration, the obtained solutions (i.e. tuple of settings) most likely differ from one another although the same objective value (i.e. deviation of distances) might have been found. Therefore, we ran groups of 20 solver calls with the same parameters to perceive the average outcome. One examination consists of a few groups of test runs which only differ in one parameter. We made examinations about changing resolutions, facets, objects, amount of events, amount of cameras and starting point. As long as there are no other assumptions the basic setup stated in Tab. \ref{tab:testsetup} is used. 
\begin{table}[!ht]
\centering
  \begin{tabular}[!ht]{|p{2.5cm} p{4.5cm}|}
	\hline
	modelled part of the scene &\ \linebreak dimension/amount\\
	\hline\hline
	surveillance area S:& cuboid $4m\times 3m\times 3m$\\
	voxel resolution: & $(16\times 12\times 12)$\\
	critical points: & all point inside the dynamic collective\\
	static collective:& 8 facets at 2 objects\\
	dynamic collective: & 24 facets at 6 objects in 2 timesteps\\
	unmodelled collective: & 24 facets at 6 objects in 3 events (of distrib.)\\
	camera placement: & 6 cameras all over the surveillance area\\
	starting solution: & cameras are placed and orientated randomly all over S\\
	stop criteria: & \mbox{maximal time limit 3h}\linebreak
			\mbox{optimization tolerance $(diagon. o. voxel)^2$}\\
	\hline
  \end{tabular}
 \caption{basic setup, which is used if no other assuptions are made}
  \label{tab:testsetup}
\end{table}

We additionally assumed that the dynamic collective is also considered to be the set of critical points. Thus, we were able to model a robot (dynamical object and critical points) spinning too fast in direction of a human (unmodelled object).

Furthermore, Tab. \ref{tab:testruns} contains all test parameters and their ranges.
\begin{table}[!ht]
\centering
  \begin{tabular}[!ht]{|p{1.75cm} p{5.25cm}|}
	\hline
	modelled part of the scene& \ \linebreak alterations\\
	\hline\hline
	voxel res.: & $(16+4i\times 12+3i\times 12+3i)$  for $i=0,1,2,3,4,5$\\
			&\\
	static coll.:& $8+60i$ facets for $i=0, \ldots,5$ at 2 obj. \\
			&\\
			&\mbox{$6+4i$ facets at $2+i$ objects $i=0, \ldots,3$}\\
			&\\
	dynamic coll.: & \mbox{$24+60i$ facets $i=0, \ldots,5$ at 3 obj./2 timest.}\\
			&\\
			 &\mbox{$2+i$ obj. $i=0, \ldots,3$ w. $24+4i$ fac./1 timest.}
				objects placed randomly\\
			&\\ 
			&\mbox{$i=1, \ldots,5$ timest. w. $2i$ obj $8i$ fac. 3 events}\\
			&\\
	unmodelled coll.: & \mbox{$24+60i$ facets $i=0, \ldots,5$ at 2 obj./3 events}\\
			&\\
			& \mbox{$2+i$ obj. $i=0, \ldots,3$ w. $24+4i$ fac./1 event}
				objects placed randomly\\
			&\\ 
			&\mbox{$i=1, \ldots,5$ events w. $2i$ obj $8i$ fac. 3 timest.}\\
			&\\
	cameras: & $i=3, \ldots,9$\\
			&\\
	restrictions of the settings' domain: 
			& \mbox{cameras placed only at `ceilling'}\linebreak
			\linebreak
			\mbox{cameras placed only in the  `upper fourth'}\\
	\hline
  \end{tabular}
 \caption{This is an overview of all examinations. An examination consists of a few groups of test runs, each group differs only in one parameter.}
  \label{tab:testruns}
\end{table}
The aim of this section is to summarize all the examinations defined in Tab. \ref{tab:testruns} and, in particular, to answer the following central questions: Can the desired optimization tolerance be satisfied in time, ie. will the target be approximated as accuratly as needed? How many iteration cycles are needed? What is the operating time of one cycle, of each iteration step and the components of one step? What is the highest memory consumption?

\subsection{Hardware and Software}

We implemented the optimization problem in C++ and compiled it with 'gcc' version 4.0.20050901 (prerelease) optimized with the setting '-O3' on SuSE Linux version 10.0.  We have used only one of the two cores of an AMD Opteron(tm) Processor 254 with 2.8 GHz Power(dynamical from 1GHz - 2.8GHz).
Further information can be taken from Tab. \ref{tab:cpuinfo} and \ref{tab:meminfo}.
\begin{table}[!ht]
\centering
  \begin{tabular}[!ht]{|p{3cm} p{4cm}|}\hline
   model name: & AMD Opteron(tm) Processor 254\\
   cpu MHz: & 1004.631\\
   cache size: & 1024kB\\
   clflush size: & 64\\
   cache\_alignment:& 64\\\hline
  \end{tabular}
 \caption{part of the output of \$: cat /proc/cpuinfo}
  \label{tab:cpuinfo}
\centering
  \begin{tabular}[!ht]{|p{3cm} p{4cm}|}
   \hline
   MemTotal: & 4038428kB\\
   MemFree: & 886856kB\\
   Buffers: & 431016\\
   Cached: & 2079360\\
   SwapCached:& 0kB\\
   Active: & 1431004kB\\
   Inactive: & 1138428kB\\
   SwapTotal: & 12586916kB\\
   SwapFree: & 12586916kB\\
   \hline
  \end{tabular}
 \caption{part of the output of \$: cat /proc/meminfo}
 \label{tab:meminfo}
\end{table}

\subsection{Optimization tolerance}

\begin{figure}[!ht]\centering
	\begin{minipage}[h]{0.85\linewidth}\centering
	\includegraphics[width=0.9\linewidth] {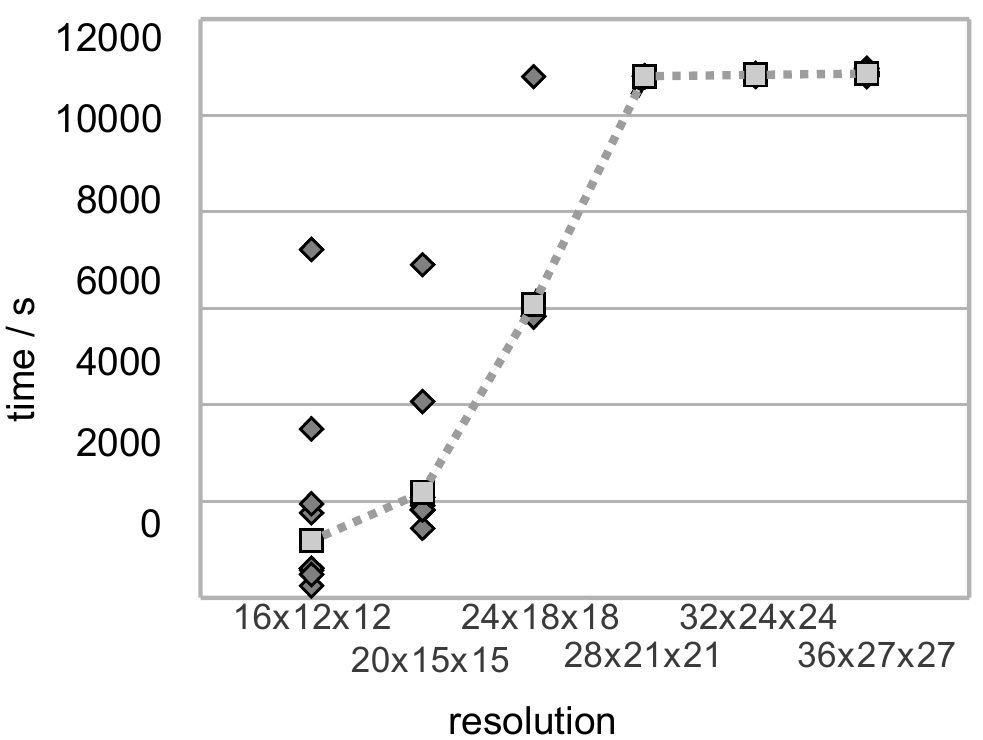}
	\caption{Scatter plot: With refined resolution the mean (light grey squares) of the amount of iteration steps (each represented in a dark grey sqare) in one group was higher. Columns: groups of 20 iterations with different resolutions;}
	\label{pic:DtimeVres}
	\end{minipage}
	\begin{minipage}[h]{0.85\linewidth}\centering
	\includegraphics[width=0.9\linewidth] {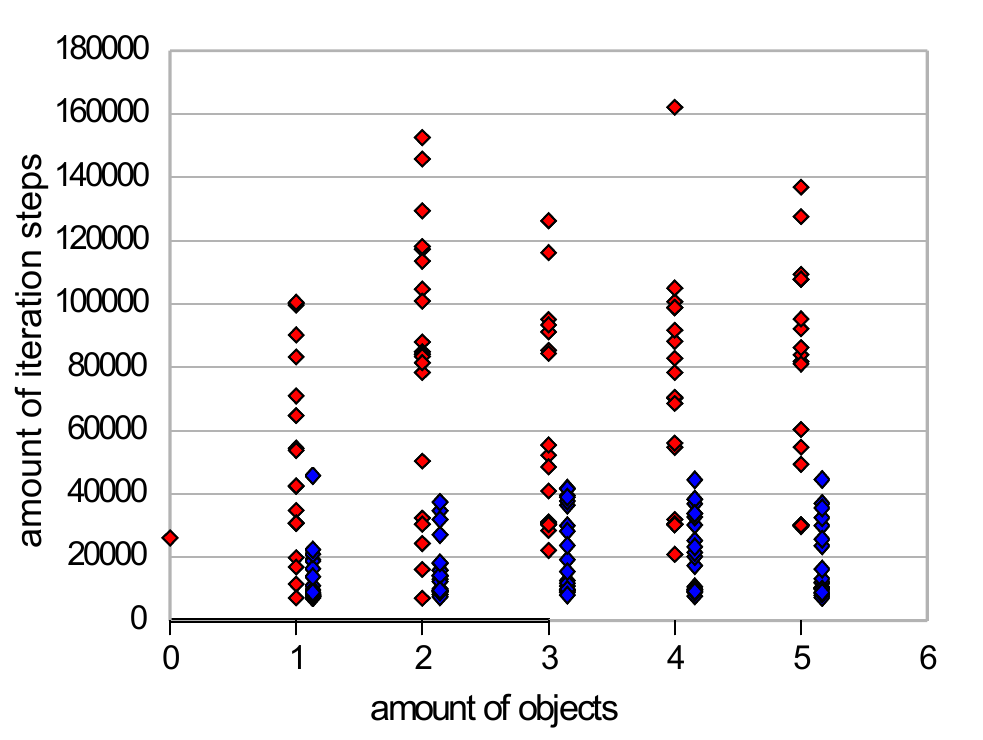}
	\caption{Scatter plot: Dynamic obstacles (and perilious points) are complicating the iteration. Columns: groups of 20 iterations of additional dynamic objects (red) and static objects (blue)}
	\label{pic:DiterVobst}
	\end{minipage}
\end{figure}

\begin{figure}[!ht]\centering
	\begin{minipage}[h]{0.46\linewidth}\centering
	\includegraphics[width=\linewidth] {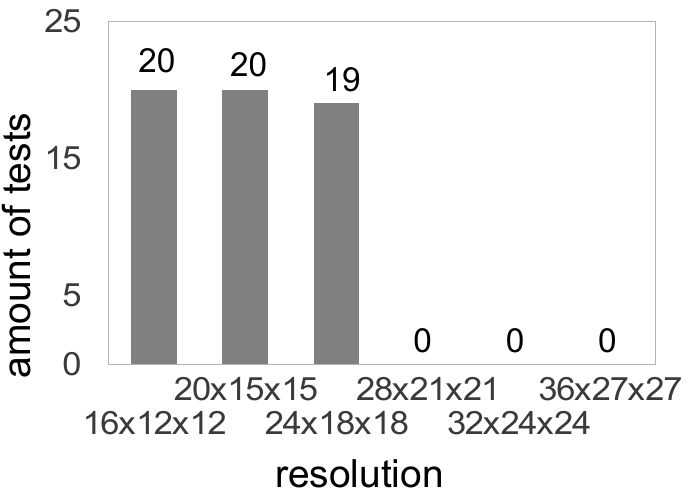}
	\caption{Bar Chart: More refined resolution than $(24\times18\times18)$ made it impossible to satisfy the predefined optimization tolerance in time}
	\label{pic:DsuccVres}
	\end{minipage} \hspace{0.3cm}
	\begin{minipage}[h]{0.46\linewidth}\centering
	\includegraphics[width=\linewidth] {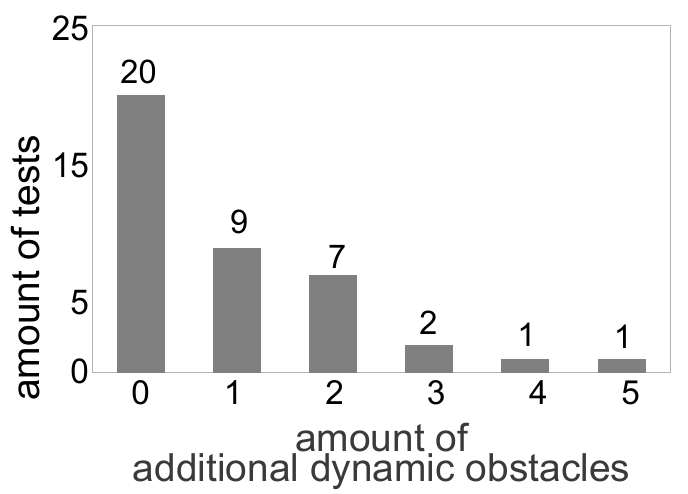}
	\caption{Bar Chart: The more dynamic objects were spread across the surveillance area, the less test runs satisfied the desired optimization tolerance}
	\label{pic:DsuccVobstdyn}
	\end{minipage}
\end{figure}

As a second stopping criteria next to the three hour time limit we introduced the optimization tolerance, which is the maximal objective value a tuple of settings must be mapped at, for the optimization to terminate. This is desinged to depend on the length of a voxel's diagonal. In many cases the solver was able to satisfy the desired optimization tolerance in the predefined maximal time. Following exceptions have exceeded the time limit: We recorded an increasing time consumption of one iteration step (beyond linear) when gradually raising the resolution of the voxel discretization. Due to the time criterion, approaches with a resolutions of more than $24\times 18\times 18$ were terminated before satisfying the optimization tolerance (see Fig. \ref{pic:DtimeVres} and \ref{pic:DsuccVres}). Increasing the number of dynamic obstacles resulted in too many iteration steps (over 160000 at most compared to less than 45000 when increasing the amount of static obstacles, cf. Fig. \ref{pic:DiterVobst}) and thus decreasing the amount of tests the optimization tolerance was satisfied for, in time, as illustrated in Fig. \ref{pic:DsuccVobstdyn}. In rare cases, a similar outcome was observed if theamount of randomly placed unmodelled objects is increased.\\
A combination of both occurrances -- the time loss in each iteration step and the requirement of too many iteration steps -- has been observed for test runs utilizing a small number of cameras (considering three cameras it was literally impossible to compute a satisfactory result, see Fig. \ref{pic:DsuccVcams}). In case of the tests on dynamic obstacles and too few cameras, the model of the unmodelled collective could not be produced optimally before the maximal computing time was up. We experienced similar results for all tests concerning restrictive domains: None of the tests reached the optimization tolerance (0.046 $m^2$) but all of them stayed below the value 0.25 $m^2$. This could be a sign, e.g. that in our test setting six cameras on the ceilling cannot assimilate the unmodelled collective close enough by the model.

\subsection{Time consumption}

\begin{figure}[!ht]\centering
	\begin{minipage}[h]{0.46\linewidth}\centering
	\includegraphics[width=\linewidth] {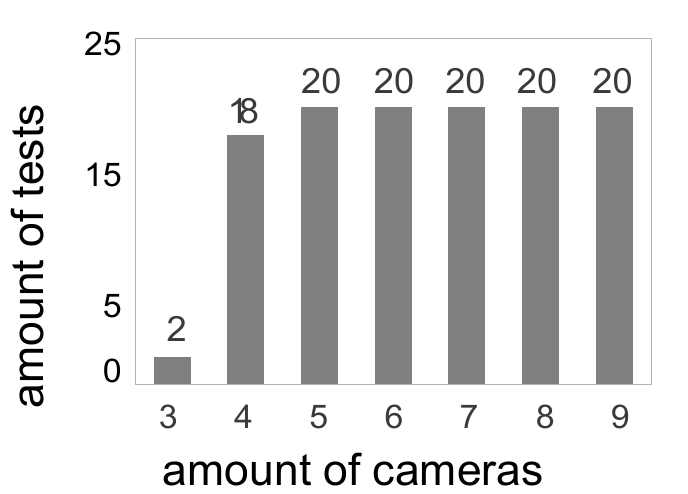}
	\caption{Bar Chart: Five till nine cameras could contour the unmodelled collective best.\newline \ \newline \ \newline}
	\label{pic:DsuccVcams}
 	\end{minipage} \hspace{0.3cm}
	\begin{minipage}[h]{0.46\linewidth}\centering
	\includegraphics[width=\linewidth] {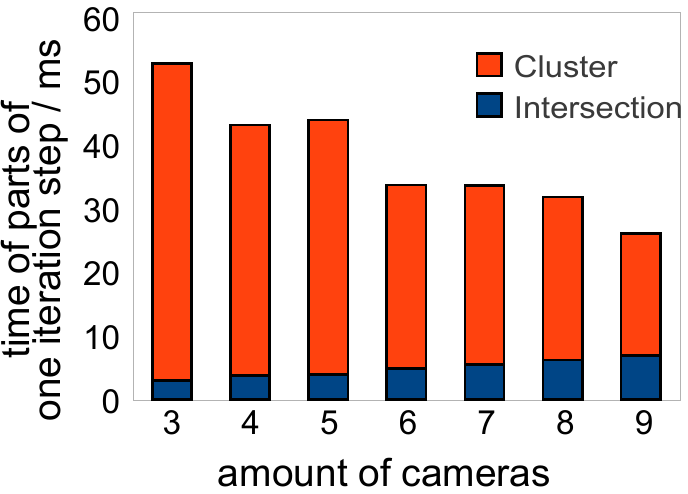}
	\caption{Bar Chart: The total time consumption of an increased amount of cameras sloped down because the clustering (orange) weighted more than the actual intersection test (blue).}
	\label{pic:DtimestepsVcams}
	\end{minipage}
\end{figure}

When raising the amount of events, time steps, facets or objects of any of the collectives we have also recorded a linearly increasing time consumption for one iteration step. Out of these, the resolution of voxel discretization and the amount of dynamic objects appear to be the most critical ones.
Using more cameras, however, resulted in a lower time loss in one iteration step in our range of camera amounts (for three cameras we required about 315 ms on average whereas for nine cameras ca. 170 ms were needed). Of course, this effect can only last until optimization tolerance is satisfied (i.e. the model assimilates the unmodelled collective as accurat as needed), and hence time consumption will slope up when using a greater amount of cameras.

Without giving a detailed explanation about the way one iteration step is calculated with our test setting's camera network, we would like to state that extending the amount of facets, cameras and refining the voxel resolution enlarges time consumption of the intersection test. However, no intersection test except for those with refined voxel resolution has exceded $15 ms$ on average. The test runs with $36\times 27\times 27$ voxel, six cameras and 24 facets have reached an average of $50 ms$. After intersecting areas the related voxels need to be combined to clusters, as to be able to check a free part's height or volume (and to compare whether it could be human). 
This task took about twice up to four times as long as the intersection test, a fact which is mainly due to its direct dependence on the resolution, but also due to the misshaping of the model (as the clustering seems to depend indirectly on the amount of cameras). As the period of an iteration step is mostly filled with intersecting and clustering, Fig. \ref{pic:DtimestepsVcams} also shows the decreasing time consumption while using more cameras.

\subsection{Memory}

\begin{figure}[!ht]\centering
	\begin{minipage}[h]{0.75\linewidth}\centering
	\includegraphics[width=0.9\linewidth] {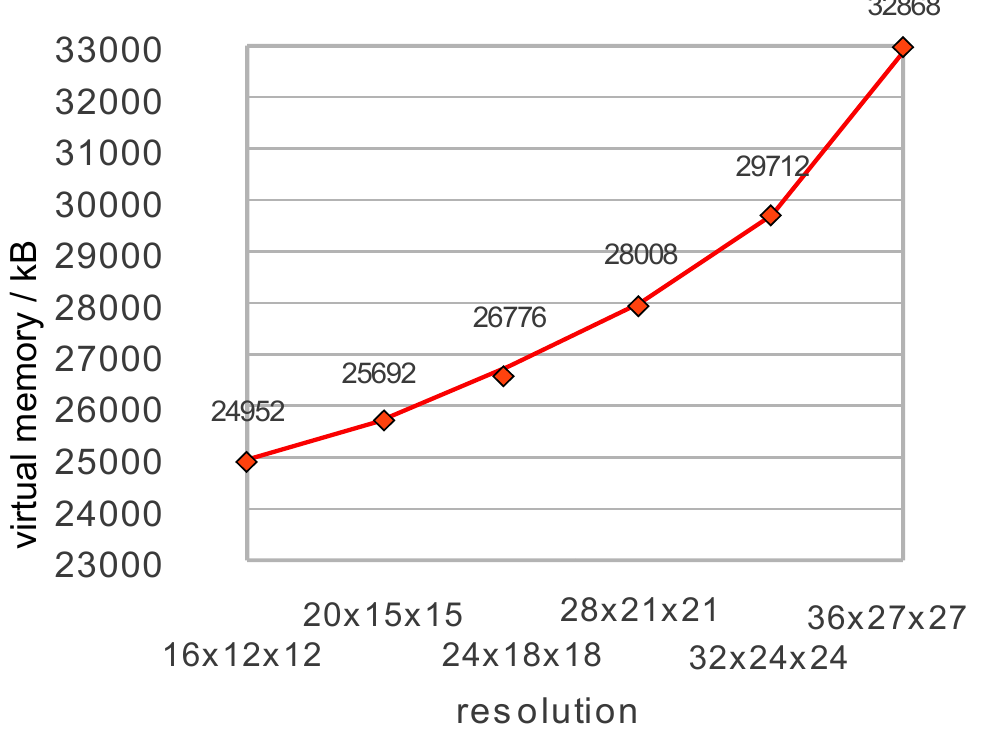}
	\caption{Plot of increasing maximal virtual memory that was used when refining the resolution of voxels}
	\label{pic:DmemVres}
	\end{minipage}
\end{figure}

Measurements of the maximal virtual memory while altering the resulution resulted an ascending graph (beyond linear), cf. Fig. \ref{pic:DmemVres}. The highest demand for virtual memory was measured while testing with the resolution $36\times 27\times 27$ (a total of 32868 kB). The graphs concerning the maximum demand for virtal memory versus facets and amount of cameras are only ascending slowly. Both show a linear slope of about 350 kB to 450 kB in our range of parameters.

\section{Conclusion and Future Prospects}
\label{sec:conclusion}

We managed to build up a camera placement optimization algorithm that computes location and orientation of a given amount of cameras inside of a specified surveillance area. Only randomly placed dynamic obstacles, too few cameras or too restricted placements and a too refined voxel resolution are a critical for this method. 
Apart from that we have succeeded to minimize the error made by evaluating distances to the visual hull of a given object up to the optimization tolerance. In contrast to existing results, we are able to model a surrounding area with static and moving obstacles without limiting camera positions or orientations and still evaluate distances conservatively.

Still, as to assimilate the model and the unknonwn collective even better, higher resolutions are desired. This leads to the fact that some improvements of the algorithm still need to be implemented. Following alterations of the algorithm may lead to an improved time consumption: First of all, it is possible to parallelize the iterations of the solver as well as some intersection tests. But as the amount of iteration steps of the solver ranged in between about $500$ and $160000$, the first goal should be to decrease both the expected number of iteration steps as well as their variance. Placing the initial position of the cameras roughly around the surveillance area and leaving the fine tuning to the algorithm could do the trick. 

Some consideration should also be paid to save many clustering and intersecting processes by leaving out unnecessary caculations. One of these calculations is the summing up $L\cdot H$ addends (the number of appearances times number of time steps), which all have to be simulated. Time loss will be minimized if cancelling the evaluation of the sum when it trespasses the current optimal value. 
Also, appropriate data structures like Oct-Trees and BSP-Trees for intersection and inclusion tests have not been implemented, yet, which improve the time loss during the intersection test.

\setbibliographyfont{name}{\textsc}
\bibliographystyle{bababbr3-lf}
\bibliography{optimal_camera_placement}

\end{document}